# Variable-sized input, character-level recurrent neural networks in lead generation: predicting close rates from raw user inputs

Giulio Giorcelli
giulio.giorcelli@gmail.com
Los Angeles, CA

*Abstract* — Predicting lead close rates is one of the most problematic tasks in the lead generation industry. In most cases, the only available data on the prospect is the self-reported information inputted by the user on the lead form and a few other data points publicly available through social media and search engine usage. All the major market niches for lead generation [1], such as insurance, health & medical and real estate, deal with life-altering decision making that no amount of data will be ever be able to describe or predict. This paper illustrates how character-level, deep long short-term memory networks can be applied to raw user inputs to help predict close rates. The output of the model is then used as an additional, highly predictive features to significantly boost performance of lead scoring models.

*Keywords* — lead scoring; deep learning; recurrent neural networks, natural language processing, character level, long short-term memory networks.

I. INTRODUCTION

Developing a powerful and reliable lead scoring model is a critical task for any lead generation company. A good lead scoring model represents the foundations of sales strategies, pricing plans, client matching as well as business development.

The amount of data collected on lead forms and from other available third party sources is often too small to model life-altering decision making such as choosing a new health insurance, finding the right legal representation or obtaining a mortgage for first time buyers. In this paper we present a variable length, character level, deep long short-term memory network that predicts lead conversions based on data points that the user has to manually type into the lead form. The exact nature of these manual inputs cannot be disclosed due to intellectual property limitation as this model is current used by the company where this model was developed.

The training process is optimized to minimize information loss and to maximize computational efficiency while the resulting model is able to find elements such as specific misspellings, overuse of punctuation and other typing patterns that are predictive of lead conversions. The output of this model is then used as an additional training input for a benchmark lead scoring model. A significant improvement in performance is observed.

Natural language processing has been previously used in the lead generation industry, however, to the best of our knowledge as one of the leading online marketing companies, this is the first time that a character level deep recurrent network has been used to predict lead conversions.

II. CORRELATION OVER ACCURACY

As explained in the previous section, due the limited amount of data available, marketing companies are usually unable to develop models that can accurately predict the outcome of each individual lead. However, lead scoring models can still be very powerful tools for identifying different quality segments of lead traffic. As shown in the hypothetical example below, even a model with low accuracy can still be highly correlated with actual close rates.

| Model score thresholds | Actual leads close rate |
|---|---|
| [0.0 - 0.2] | 0.5% |
| [0.2 - 0.4] | 1.2% |
| [0.4 - 0.6] | 2.0% |
| [0.6 - 0.8] | 2.8% |
| [0.8 - 1.0] | 4.0% |

**Table 1**: a hypothetical example that shows how a model with low accuracy can be very effective at differentiating quality segments of lead traffic.

Consequently, metrics such as accuracy, F-1 score and recall are almost meaningless in evaluating these models. We therefore use the Pearson two-tailed correlation coefficient – in short, Pearson's r:

$$r = \frac{\sum XY - \frac{\sum X \sum Y}{N}}{\sqrt{\left(\sum X^2 - \frac{(\sum X)^2}{N}\right)\left(\sum Y^2 - \frac{(\sum Y)^2}{N}\right)}}$$

**Fig. 1**: formula for Pearson two-tailed correlation coefficient

## III. MODEL ARCHITECTURE

The model consists in four long short-term memory layers [2] (LSTM in short) with 256 units each. Each layer is followed by a dropout layer [3] with 30% probability. The stacked output of the LSTM layers is then passed through a fully connected layer which consequently feeds its output to a sigmoid function.

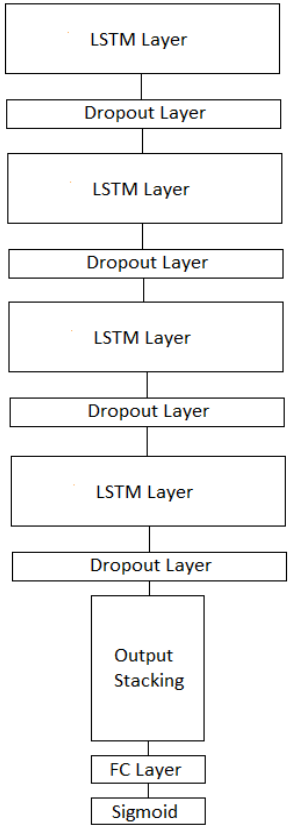

**Fig. 2:** architecture of the network used for this project

The peculiarity of this model is the absence of an embedding layer [4]. Embedding layers have been shown to improve performance of LSTM networks [4]. In this use case, however, our model achieves a higher correlation coefficient on test data and it also converges faster without an embedding layer at all, regardless of whether pertained or untrained weights were being used.

| Embedding Layer | Pearson's R | P-value |
|---|---|---|
| Present, untrained | 0.05 | 0.02 |
| Present, pre-trained | 0.08 | < 0.01 |
| None | 0.13 | < 0.01 |

**Table 2**: removing the embedding layer yields better performance

## IV. TRAINING THE MODEL

The model is trained with the Adam optimization method [5] at a learning rate of 0.001 for the first 55 epochs and with stochastic gradient descent [6] (SGD in short) with Nesterov momentum [8] for the last 20 epochs at a learning rate of 0.002. Switching from Adam to SGD for the final epochs of training is a proven way to improve generalization in deep LSTM networks [7].

The model is trained without the use of fixed-length inputs, but with variable-sized mini-batches of 64 samples each. The lack of fixed-length inputs has two main advantages:

1. Increased computational efficiency
2. Minimized information loss

### A. Increased computational efficiency

Pared with fixed-sized batches, padding often results in large sparse matrices that are often only minimally utilized. For example, the median length of the strings in the training dataset used for this project is 68, while the longest string is 214 characters long. A happy medium for a hypothetical fixed-sized batch should be between 120 and 160 elements. This would resulte in the majority of the matrix being empty, forcing a useless computational cost on our system. Using variable-sized mini-batch reduces this issue to a minimum.

| Y | E | L | L | O | W | Ø | Ø | Ø | Ø |
|---|---|---|---|---|---|---|---|---|---|
| B | L | U | E | Ø | Ø | Ø | Ø | Ø | Ø |
| R | E | D | Ø | Ø | Ø | Ø | Ø | Ø | Ø |

**Fig. 3.1**: example of mini-batch with padding. Almost half of the matrix is empty.

| Y | E | L | L | O | W |
|---|---|---|---|---|---|
| B | L | U | E | Ø | Ø |
| R | E | D | Ø | Ø | Ø |

**Fig. 3.2**: the same data represented in a variable-sized mini-batch. Notice how most of the matrix is now utilized.

### B. Minimized information loss

Using fixed-sized batches means that the maximum string length has to be arbitrarily decided before training the model. All the strings that are longer than the maximum length parameter will be truncated, resulting in severe information loss. Variable-sized mini-batches solve this issue by using batches that mirror the size of the longest string in the batch.

For testing purposes a model was trained using fixed-sized batches and another model was trained on the same dataset with variable-sized batches. The latter converged faster and obtained a higher correlation coefficient on test data:

| Batch size | Pearson's R | P-value | Epochs to Convergence |
|---|---|---|---|
| Fixed | 0.11 | <0.01 | 92 |
| Variable | 0.13 | <0.01 | 74 |

**Table 3**: the model trained with variable-sized batches converges faster and yealds higher correlation on test data

## V. RNN's OUTPUT AS AN INPUT FEATURE

The last step of this projects consists in understating whether the information extracted from manually typed-in inputs by the newly trained character-level RNN can help predict lead closes.

A benchmark lead scoring model is trained on a number of features captured throughout the lead form except all the inputs that are manually typed in by the user. The exact nature of these features cannot be revealed as they are currently in use for commercial applications.

A second lead scoring model is trained including the output of the character-lever RNN on users' manual inputs as an additional feature. Except for the aforementioned difference, both models, including their training, validation and test datasets are identical.

The model that includes the RNN output as an additional input feature is able to train for longer without overfitting achieving lower validation loss and it yields higher correlation on test data.

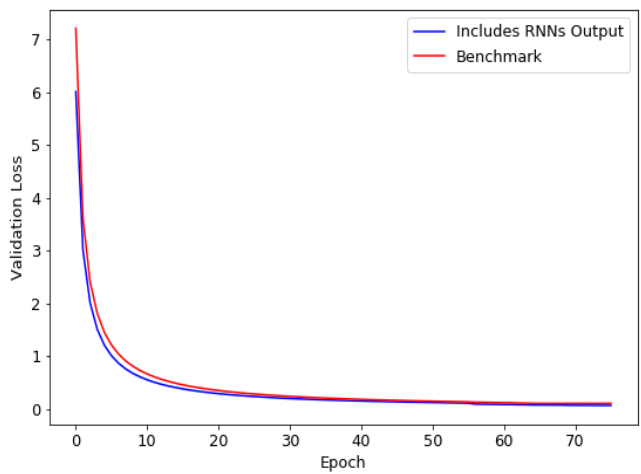

**Fig. 4**: comparison of validation losses for benchmark model and model trained including char-RNN's output as an additional input feature.

| Model | Pearson's R | P-value |
|---|---|---|
| Benchmark | 0.21 | <0.01 |
| Includes RNN's Output | 0.32 | <0.01 |

**Table 4**: including the char-RNN's output as an additional input feature dramatically increases performance

## CONCLUSION

Lead scoring plays an integral part in the optimization and business development strategies of nearly every marketing company. Extracting every bit of value from data collected on the lead submitter is key to succeeding at those strategies. With this variable-sized input, character level LSTM network we find a way extract a sizeable amount of information from inputs that are manually typed in by the lead submitter. These type of fields, to best of our knowledge, has been historically undervalued if not completely unutilized in lead scoring. Using the model illustrated in this paper to extract information from such features and then feeding its output to a lead scoring model dramatically improve the performance of the latter, resulting in a major advantage for the business.